\documentclass[10pt,twocolumn,letterpaper]{article}

\usepackage{wacv}
\usepackage{times}
\usepackage{epsfig}
\usepackage{graphicx}
\usepackage{amsmath}
\usepackage{amssymb}
\usepackage{booktabs}

\usepackage{xcolor}
\usepackage{epsfig}
\usepackage{graphicx}
\usepackage{amsmath}
\usepackage{enumitem}

\usepackage[labelfont=bf]{caption}

\usepackage{physics}
\usepackage{graphicx}
\usepackage{float}
\usepackage{amsmath}
\usepackage{subfigure}
\usepackage{amsmath}
\usepackage{longtable}
\usepackage{booktabs}
\usepackage{float}
\usepackage{threeparttable}
\usepackage{xcolor}
\usepackage{diagbox}
\usepackage{hyperref}

\usepackage{amsfonts}
\usepackage{amsmath}
\usepackage{amssymb}

\usepackage{multirow}
\usepackage{multicol}
\usepackage{tabularx}
\usepackage{dsfont}
\usepackage{url}

\usepackage{amsthm}
\usepackage[export]{adjustbox}
\usepackage{comment}
\usepackage{cite}
\usepackage{hyperref}

\usepackage[belowskip=1pt,aboveskip=0pt]{caption}

\usepackage[utf8]{inputenc} 
\usepackage[T1]{fontenc}    
\usepackage{url}            
\usepackage{booktabs}       
\usepackage{amsfonts}       
\usepackage{nicefrac}       
\usepackage{microtype}      

\definecolor{dg}{rgb}{0,0.694,0.298}
\definecolor{purple}{rgb}{0.4,0.176,0.569}

\usepackage{pifont}
\usepackage{xspace}
\makeatletter
\DeclareRobustCommand\onedot{\futurelet\@let@token\@onedot}
\def\@onedot{\ifx\@let@token.\else.\null\fi\xspace}

\makeatother

\def\wacvPaperID{324} 

\wacvfinalcopy 

\begin{document}

\title{Pik-Fix: Restoring and Colorizing Old Photos}



\author{Runsheng~Xu$^1$\thanks{Equal contribution. $^\dagger$ Corresponding author.},~Zhengzhong~Tu$^{2*}$,~Yuanqi~Du$^{3*}$,~Xiaoyu~Dong$^{4}$,~Jinlong~Li$^5$,~Zibo~Meng$^6$\\Jiaqi~Ma$^1$,~Alan~Bovik$^2$,~Hongkai~Yu$^{5\dagger}$
\\
$^1$~University of California, Los Angeles,~$^2$~University of Texas at Austin,~$^3$~Cornell University\\ $^4$~Northwestern University, $^5$~Cleveland State University, $^6$~Innopeak Technology Inc.\\
{\tt\small rxx3386@ucla.edu, hongkaiyu2012@gmail.com}
}


\maketitle
\thispagestyle{empty}

\begin{abstract}
Restoring and inpainting the visual memories that are present, but often impaired, in old photos remains an intriguing but unsolved research topic. Decades-old photos often suffer from severe and commingled degradation such as cracks, defocus, and color-fading, which are difficult to treat individually and harder to repair when they interact. Deep learning presents a plausible avenue, but the lack of large-scale datasets of old photos makes addressing this restoration task very challenging. Here we present a novel reference-based end-to-end learning framework that is able to both repair and colorize old, degraded pictures. Our proposed framework consists of three modules: a restoration sub-network that conducts restoration from degradations, a similarity network that performs color histogram matching and color transfer, and a colorization subnet that learns to predict the chroma elements of images conditioned on chromatic reference signals.
The overall system makes uses of color histogram priors from reference images, which greatly reduces the need for large-scale training data.
We have also created a first-of-a-kind public dataset of real old photos that are paired with ground truth ``pristine'' photos that have been manually restored by PhotoShop experts. We conducted extensive experiments on this dataset and synthetic datasets, and found that our method significantly outperforms previous state-of-the-art models using both qualitative comparisons and quantitative measurements.
The code is available at \url{https://github.com/DerrickXuNu/Pik-Fix}.
\end{abstract}

\begin{figure}[!ht]
\centering
\footnotesize
  \includegraphics[width=0.95\columnwidth]{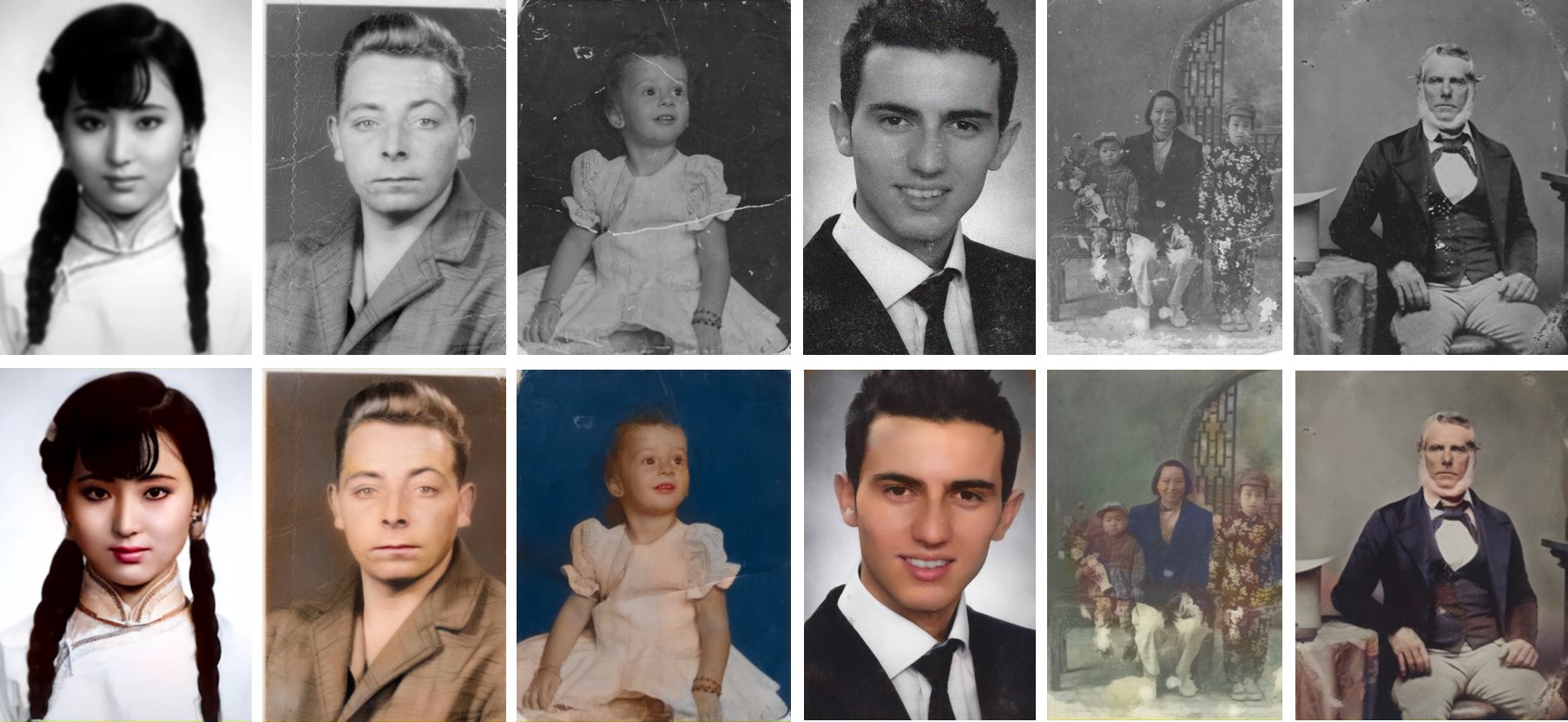}
  \caption{Examples of old photo repair~(restoration and colorization) generated by Pik-Fix. \textmd{Pik-Fix is able of simultaneously repair multiple image degradations of a photograph while also colorizing it.}
  }
  \label{fig:firstresult}
  \vspace{-5mm}
\end{figure}


\section{Introduction}
\label{sec:intro}

While our experience of the visual world are colorful, in earlier days of photography pictures were usually captured as ``black and white'', i.e. as gray-scale. As time elapses, they suffer other degradation as well. While consumer service are available for restoring and colorizing old photos, these require significant expertise in image manipulation, which is labour intensive, costly, and time-consuming. Thus, developing automated systems that can rapidly and accurately colorize and restore old photos is of interest.

Recently, deep learning based techniques have achieved high performance levels on a broad range of computer vision problems~\cite{zhang2016colorful,he2018deep,xu2022v2x,zhao2022tracking,xu2022opv2v,tu2022maxvit,fan2022object,xu2022cobevt, paz2022tridentnetv2, paz2020probabilistic,chen2022model,2022arXiv220913679X,xu2021opencda}. They have also been successfully applied to image restoration such as image denoising~\cite{chen2018learning, meng2020gia}, super-resolution~\cite{ledig2017photo, mei2020dsr}, deblurring~\cite{nah2017deep,tu2022maxim}, colorization~\cite{zhang2016colorful,he2018deep}, and compression~\cite{balle2016end,chen2020proxiqa}.
However, learning-based colorization models generally require large-scale training datasets ~\cite{zhang2016colorful} to obtain favorable performance, which is energy-inefficient, labor-intensive, and time-consuming. Towards reducing large data requirements, the authors of \cite{ironi2005colorization,gupta2012image,he2018deep, zhang2019deep} proposed to employ reference/example images to assist colorization of gray-scale images. He \textit{et al.}~\cite{he2018deep} uses separate similarity and colorization networks. However, since there are inherent ambiguities of the colors of natural objects because of the effects of ambient lighting. Better results than pixel-level color matching may be obtained by deriving features that describe the statistical color distributions of the reference pictures. In this direction, Yoo~\cite{yoo_fewshot} deploy the means and variances of deep color features, but do not utilize second-order~(spatial) distribution models, thereby  discarding information descriptive of correlations that exist within image textures and their colors.

Since the spatial statistical color distribution is a very likely a useful source of colorization features, we have developed a reference-based, multi-scale  spatial color histogram fusion method of image colorization. Using reference pictures to guide the colorization of gray-scale photographs relieves the need for large-scale training data. Precisely, we devised a novel end-to-end deep learning framework for old photo restoration which we dub Pik-Fix, which is composed of 1) a convolutional sub-network that is trained to conduct degradation restoration, 2) a similarity sub-network that performs reference color matching, and 3) a colorization sub-network that learns to render the final colorful image. As illustrated in  Fig.~\ref{fig:firstresult}, Pik-fix can restore and colorize the old degraded photos using only limited training data, making it attractive for data-efficient applications. Previous methods~\cite{wan2020bringing} mainly use the quantitative results on synthetic data with the restoration ground truth and qualitative results on collected real data without the restoration ground truth for experimental evaluations. To the best of our knowledge, there exists no similar public dataset of authentic, real-world degraded and gray-scale photos that are associated with pristine reference versions of the same photos.  Towards advancing research in this direction, we designed and built \textit{a first-of-a-kind real-world old photo dataset} consisting  of 200 authentic old grayscale photos, where each old photo is paired with a `pristine' version of it that was manually restored and colorized by Adobe Photoshop editors. Our experimental results show that Pik-Fix can outperform  state-of-the-art methods on both existing public synthetic  datasets and on our real-world old photo datasets, even though it requires much less training data.
Our major contributions are summarized as follows:

\vspace{-2pt}
\begin{itemize}[leftmargin=*,noitemsep]
    \item We propose \textit{the first end-to-end deep learning framework} (Pik-Fix) that learns to simultaneously restore and colorize old photos, only requiring a small amount of training data.
    \item A \textit{reference-based multi-scale  color histogram fusion method} for image colorization that learns the content-aware transfer functions between the input and reference.
    \item \textit{The first publicly available  dataset of authentic, real-world degraded old photographs}. Each of these 200 authentic contents is paired with a 'pristine' version that were manually restored and colorized by Photoshop editors.
    \item Our experimental results show that the model, called Pik-Fix, achieves better visual and numerical performance than state-of-the-art methods on existing synthetic data and on our new real-world dataset.
\end{itemize}


\begin{figure*}[!t]
\centering
\includegraphics[width=0.9\textwidth]{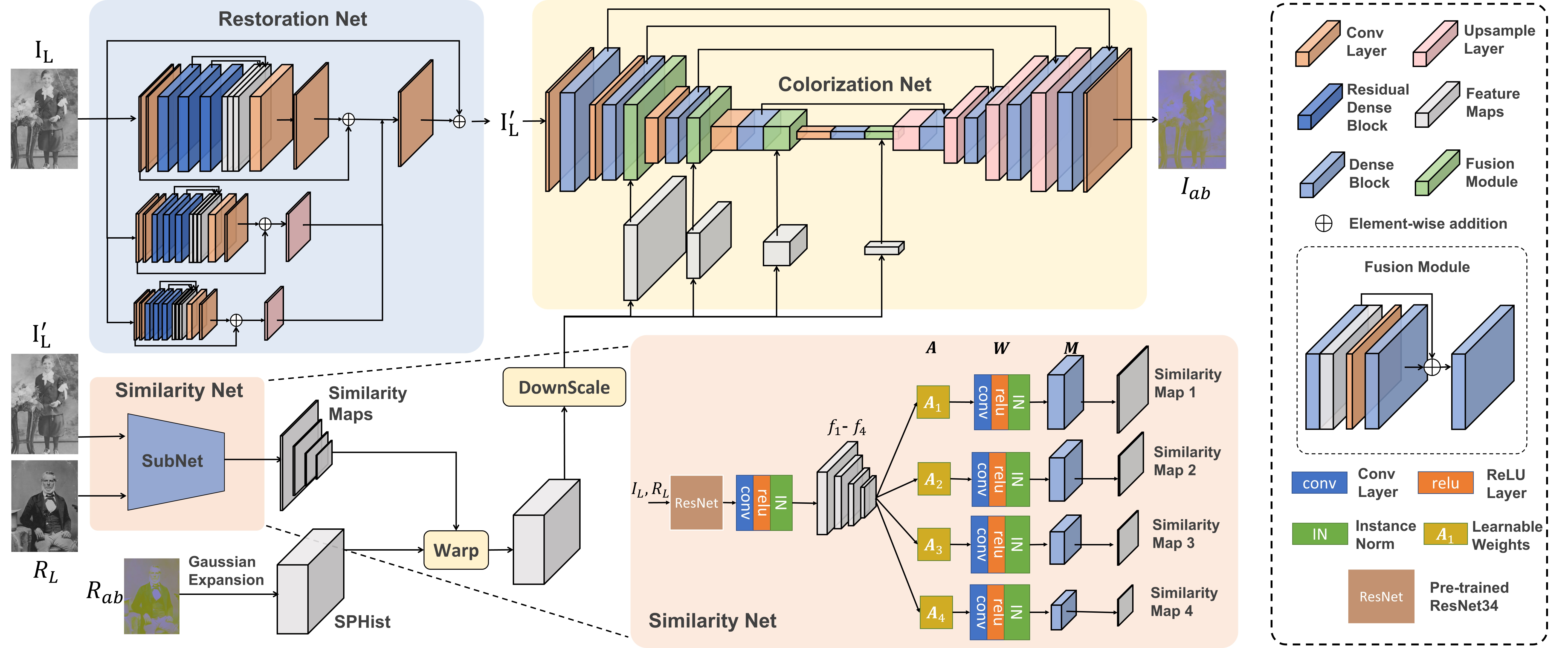}
\caption{Flow diagram of the triplet networks (restoration, similarity and colorization) that define the flow of visual information processing in Pik-Fix.}
\label{fig:overview}
\vspace{-3mm}
\end{figure*}

\section{Related Work}
\label{sec:related-work}

\subsection{Image Colorization}

Driven by deep neural networks, automatic image colorization have made great progress recently~\cite{zhao2018pixel,deshpande2015learning,cheng2015deep,zhang2017real}.  Semantics analysis has been identified for successful colorization. For example,  \cite{iizuka2016let} and \cite{zhao2018pixel} design two-branch architectures that explicitly learn to fuse the local image features with global semantic predictions. The authors of  \cite{su2020instance} argue that pixel-level analysis is insufficient to learn subtle variations of  object appearance and color, and shows that incorporating object-level analysis into that regression architecture yields better performance. Some works also try to employ reference images to help colorization and use a variety of ways to compute correspondence between the input pictures and the reference data, including pixel comparison \cite{welsh2002transferring,liu2008intrinsic}, semantic matching \cite{ironi2005colorization,charpiat2008automatic}, and super-pixel level \cite{gupta2012image,chia2011semantic} similarities.

\subsection{Image Restoration}
There is a wide array of degradations that can affect older photographs, including some that occurred during capture, such as film grain and blur, and others that occur over time, like stains, fading color, and cracks.  Traditional computational approaches to restore photos that have been digitized usually involve the application of prior constraints such as non-local self-similarity \cite{buades2005non}, sparsity \cite{elad2006image}, or local smoothness \cite{weiss2007makes}.
More recently, deep learning-based methods have proved efficacious on many picture restoration  tasks, such as image denoising \cite{zhang2017learning,zhang2017beyond,zhang2018ffdnet},  super-resolution \cite{dong2014learning,kim2016accurate,ledig2017photo}, and deblurring \cite{xu2014deep,sun2015learning,nah2017deep}.  The success of these methods derives from the ability to simultaneously learn smooth semantics, and perceptual and local image representations.


\subsection{Old Photo Restoration.}
Old Photo Restoration aims at removing the degradations of old photos and colorizing them with natural colors. However, most of the existing models only address one particular aspect of old photo restoration, color restoration, or degradation restoration. The authors \cite{wang2018high} designed an image-level pixel-to-pixel image translation framework using paired synthetic and real images. A model called Deoldify~\cite{jantic} also implements a pixel-to-pixel translation using a  GAN. \cite{ulyanov2018deep} learns to conduct single-degradation image restoration in an unsupervised manner. \cite{wan2020bringing} first encodes image data into latent representations that separate old photos, ground truth, and synthetic images. It learns image restoration by producing the latent translation.

Although previous have been able to deliver perceptual equality by solely conducting colorization or restoration, in most instances old photo restoration requires both colorization and distortion restoration. Our work leverages both learning-based restoration and example-based color restoration methods to obtain old photo restoration that addresses both aspects. Importantly, our example-based colorization technique requires much less training data.

\section{Methodology}
\label{sec:methodology}
There are several major challenges that need to be addressed to further advance old photo restoration. Complex, commingled degradations are often observed in real-world old photos, which are impossible to model analytically, and difficult to gather into large amounts of representative training data. Further, colorization is an ill-posed, ambiguous problem \cite{Charpiat2008}, hence existing models require very large training datasets.
The presence of complex distortions can make colorization harder. While this has not been deeply studied, the loss of real information likely impedes inferencing and regression. Conversely, restoring degraded gray-scale photos may be harder without clues supplied by color, which tends to be smooth and regional. Solving both problems together has the potential to improve the overall solution.

Towards overcoming these challenges, we propose an end-to-end framework, as depicted in Fig.~\ref{fig:overview}.
Denoting the input grayscale photo as $I_{L} \in \mathbb{R}^{H\times W \times 1}$, the restoration sub-net attempts to reverse any degradations to produce a restored gray-scale image $I_{L}^{'} \in \mathbb{R}^{H\times W \times 1}$. Then, $I_{L}^{'}$ and the luminance channel of an associated reference picture $R_L\in \mathbb{R}^{H\times W \times 1}$ are both fed into the similarity sub-net, which produces a similarity map. Then, the chromatic features from the $ab$ channels $R_{ab} \in \mathbb{R}^{H\times W \times 2 }$ of the reference image are projected onto the input image space. The colorization sub-net accepts $I_{L}^{'}$ and the projected reference color features together as inputs, processing them to generate $ab$ channels $I_{ab} \in \mathbb{R}^{H\times W \times 2}$, finally concatenating it with $I_{L}^{'}$  to obtain a restored and colorized result $I_{Lab} \in \mathbb{R}^{H\times W \times 3}$.

While previous methods operate by directly feeding the raw $ab$ channels of the reference image into a colorization network~\cite{he2018deep, zhang2019deep}, or utilize low-order statistics (\textit{e.g.}, mean and variance) of adaptive instance normalization of the reference image~\cite{ Xu_2020_CVPR, yoo_fewshot}, we instead employ a multi-scale fusion method that combines a spatial-preserving color histogram with deep features. The spatial-preserving color histogram contains useful prior information regarding the spatial relationships of color. The color features and deep features are aggregated over multiple scales, enabling the learning of the colorization process without a large number of training samples.
In the following sections, we detail the restoration sub-net, similarity sub-net, colorization sub-net, and reference selection algorithms.

\subsection{Restoration Sub-Net}
Broadly, the types of degradations that affect old photos can be divided into two categories: physical defects (\textit{e.g.}, cracks, tears, smudges) and capture defects (\textit{e.g.}, blur, exposure)~\cite{wan2020bringing}. Correcting physical defects typically requires that the receptive fields of the analyzing neural network be large enough to capture impairments that span much of the photo dimensions. Yet it is also important that the network accesses local information since capture distortions usually manifect locally, even when globally present.

Here we address the bifurcated nature of old photo distortions by developing a multi-level Residual Dense Network (RDN~\cite{rdn}) that serves as the restortion sub-net. RDN models have previously demonstrated outstanding performance on common image restoration tasks like super-resolution, denoising, and deblurring, mainly facilitated by a core module called the residual dense block. The residual dense block is able to extract abundant information via the use of dense connections and contiguous memory mechanisms. While the RDN architecture has been shown to be suitable for handling capture defects, it processes images at a single resolution, restricting the sizes of the filter receptive fields and weakening its ability to correct physical flaws. To enable RDN to handle the broader range of distortions, we have formulated a multi-level RDN that is able to analyze distorted pictures over an enlarged span of receptive field sizes. 

As shown in Fig.~\ref{fig:overview}, 
an original picture, along with 4$\times$ and 8$\times$ downsampled versions of it are fed into the top, second, and third levels of the RDN, respectively. Each level consists of three residual dense blocks, each composed of 4 identical residual dense units. The outputs of the lower levels are upsampled via bilinear interpolation and fused via concatenation, then passed through another convolution layer to generate the restored luminance $I_{L}^{'} $.

\subsection{Similarity Sub-Net}

After the refined luminance map  $I_{L}^{'}$ is obtained from the restoration sub-net, it is passed to the similarity sub-net along with the reference image's luminance channel $R_{L}$. The similarity sub-net is designed to project the reference image features onto the feature space of the input picture. As illustrated in Fig.~\ref{fig:overview}, a pre-trained ResNet34~\cite{resnet} is employed to retrieve layer1, layer2, layer3, layer4 feature maps from the input and reference pictures, respectively. Note that these feature maps have progressively smaller spatial resolutions and a larger number of feature channels with increased network depth. Then, four convolution layers are applied to these intermediate features, yielding feature maps having the same channel dimensions  $f_i \in  \mathbb{R}^{H_i \times W_i \times C}$ ($i=1,2,3,4$). We utilize similarity maps at multiple scales to later allow for multi-level feature fusion in the colorization sub-net. Rather than simply resizing and concatenating the four feature maps, we propose to construct a learnable coefficient $A_i \in  \mathbb{R}^{1\times4}$, where $i=$1 to 4, that assigns different weights to the feature maps depending on the target similarity map size. These weighted feature maps are then concatenated together to obtain a feature tensor. For instance, the concatenated feature  $M_i\in  \mathbb{R}^{H_i \times W_i \times C}$ at scale $i$ would be:
\begin{equation}
\begin{split}
     M_i =  W \circledast [ &\text{g}(A_{i1} * f_1) \oplus \text{g}(A_{i2} * f_2) \\
     \oplus
     & \text{g}(A_{i3} * f_3) \oplus \text{g}(A_{i4} * f_4) ],
\end{split}
\end{equation}
where $W$ is the shared convolution filter for the convolution operator $\circledast$, \text{g}  is an up-sampling or down-sampling function that aligns the feature size to the target similarity map size, $*$ indicates element-wise multiplication, and $\oplus$ denotes the concatenation operation. Subsequently,  the three-dimensional feature vector $M_i$ will be reshaped to a two-dimension matrix $\overline{M_i} \in \mathbb{R}^{H_i W_i \times C} $. Then, the similarity map $ \Phi_{R\leftrightarrow I}^{i}\in  \mathbb{R}^{H_i W_i \times H_iW_i}$ characterizing the correlation structure between the reference picture $R$ and the input picture $I$ at scale level $i$ is computed at each spatial location $(u,v)$ as follows:
 \begin{align}
 \resizebox{0.88\hsize}{!}{%
     $\Phi_{R\leftrightarrow I}^{i}(u,v) = \frac{(\bar{M_i}^I(u) - \mu_{\bar{M_i}^I})\cdot(\bar{M_i}^R(v) - \mu_{\bar{M_i}^R})}{||\bar{M_i}^I(u) - \mu_{\bar{M_i}}||_2 ||\bar{M_i}^R(u) - \mu_{\bar{M_i}^R}||_2 },$%
     }
 \end{align}
 where $\mu_{\bar{M_i}^I} $ and $\mu_{\bar{M_i}^R}$ are mean feature vectors. The softmax function is then applied to the elements of the similarity map along the x-axis so each mapped element lies within [0,1]. This similarity map is then passed to the colorization sub-net, whose task is simplified since the reference picture's information is aligned with that of the input image.


\subsection{Colorization Sub-Net}

To tackle the aforementioned colorization problem, we develop a method of guiding the process using the color prior in the reference picture.
Specifically, we utilize a space-preserving color histogram (SPHist) computed on the reference pictures. Unlike the traditional color histogram, the SPHist can retain spatial picture information while modeling the probability that each pixel color falls within each bin. Importantly, \textit{SPHist is differentiable}, and thus, it can be used in an end-to-end neural network trained using gradient back-propagation. We accomplish this by using Gaussian expansion~\cite{schutt2017_gaussian} to separately approximate the SPHist $h \in \mathbb{R}^{H \times W \times K}$ of each channel, where $K$ is the number of histogram bins. Then, the probability of a pixel at location ($i,j$) falling into the $k$-th bin is expressed as follows:
\begin{equation}
    \label{eq:histogra}
    h(i,j,k) = \frac{\text{exp}(-{(D_{ij}-u_k)^2}/{2\sigma^2})}{\sum_{k=1}^{K}\text{exp}(-{(D_{ij}-u_k)^2}/{2\sigma^2})},
\end{equation}
where $D_{ij}$ is the value of $a$ (or $b$) channel of the reference picture at spatial coordinate ($i,j$); the spread of the Gaussian distribution is fixed at $\sigma=0.1$; $u_k$ is a learnable parameter representing the center of bin $k$, which is initialized as:
\begin{align}
    u_k^0 = v_{min} + ({v_{max} - v_{min}})/{K} * k, 
\end{align}
where $v_{min}$ and $v_{max}$ are the minimum and maximum possible values of the $ab$ channels (-1 and 1, respectively in our experiments). Although the bins are equally distributed at the start, after training over several iterations, their distributions become unequal, since some colors are rarer than others `in the wild'. The extracted color histogram is reshaped to  $\overline{h} \in \mathbb{R}^{HW \times K} $ and down-sampled to the available four scales to enable matrix multiplication with the corresponding scales of similarity maps, leading to a warped SPHist that contains similarity-guided space-preserved color histogram from the reference picture. The warped SPHist is then fed into different levels of the encoder in the colorization sub-network to conduct color prediction.

The backbone of our colorization network employs a global U-Net shape~\cite{unet} with densenet blocks~\cite{denseblock}. There are four dense blocks in the encoder containing 6, 12, 24, and 16 dense units. The decoder shares a similar structure as the encoder, and bi-linear interpolation is employed to upscale the forwarding features between the dense blocks. The warped SPHist extracted from the reference picture is concatenated with the intermediate features after each dense block in the encoder, yielding inputs to the fusion module. The fusion module contains a dense block with six dense units and a $3*3$ convolution layer, which is responsible for efficiently combining the traditional color heuristics and deep features to enable accurate colorization. Since the reference information of reference is fused during the intermediate stages instead of at the start, the model learns to deal with dissimilarities between the input and reference pictures in a multi-scale manner.

\begin{figure*}[!t]
\centering
\def\xlinewidth{0.135}
\def\ylinewidth{0.09}
\def\xem{-1pt}
\def\yem{1pt}
\setlength{\tabcolsep}{1.2pt}
\begin{tabular}{ccccccc}

\includegraphics[ width=\xlinewidth\linewidth, height=\ylinewidth\linewidth]{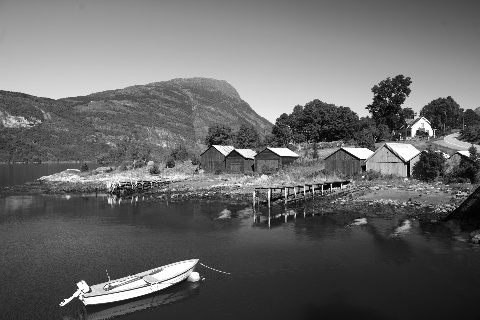} &
\includegraphics[ width=\xlinewidth\linewidth, height=\ylinewidth\linewidth]{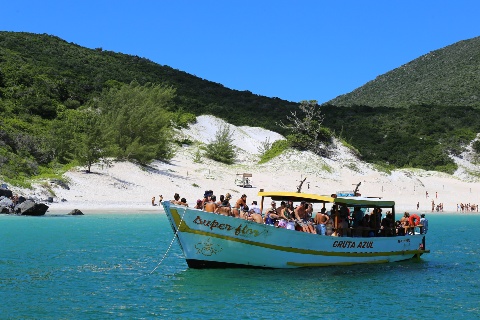} &
\includegraphics[ width=\xlinewidth\linewidth, height=\ylinewidth\linewidth]{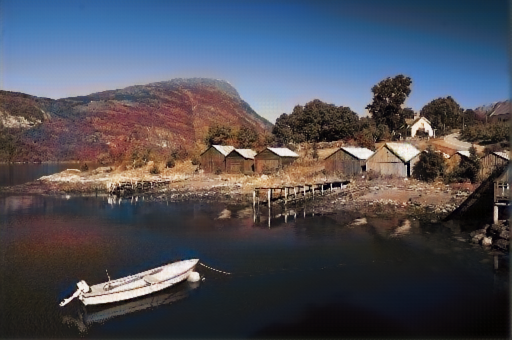} &
\includegraphics[ width=\xlinewidth\linewidth, height=\ylinewidth\linewidth]{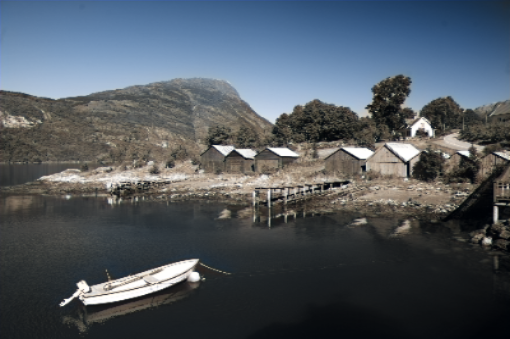} &
\includegraphics[ width=\xlinewidth\linewidth, height=\ylinewidth\linewidth]{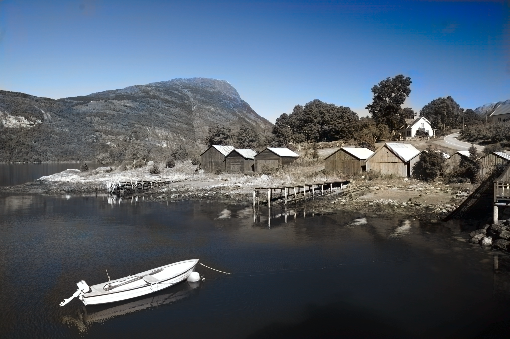} &
\includegraphics[ width=\xlinewidth\linewidth, height=\ylinewidth\linewidth]{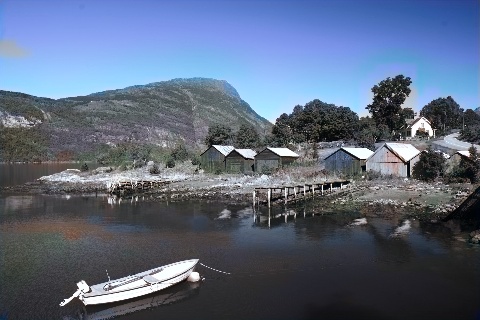} &
\includegraphics[ width=\xlinewidth\linewidth, height=\ylinewidth\linewidth]{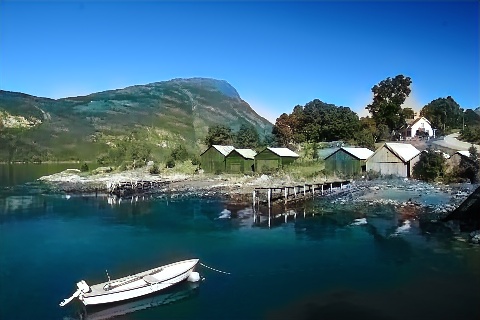} \\[\xem]

\includegraphics[ width=\xlinewidth\linewidth, keepaspectratio]{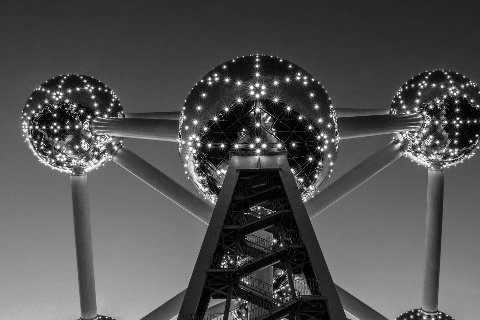} &
\includegraphics[ width=\xlinewidth\linewidth]{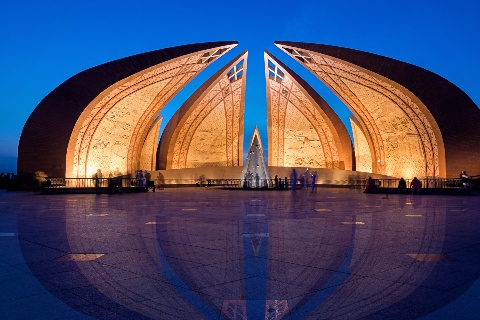} &
\includegraphics[ width=\xlinewidth\linewidth, height=\ylinewidth\linewidth]{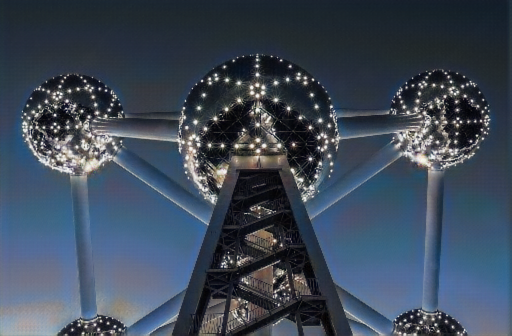} &
\includegraphics[ width=\xlinewidth\linewidth, height=\ylinewidth\linewidth]{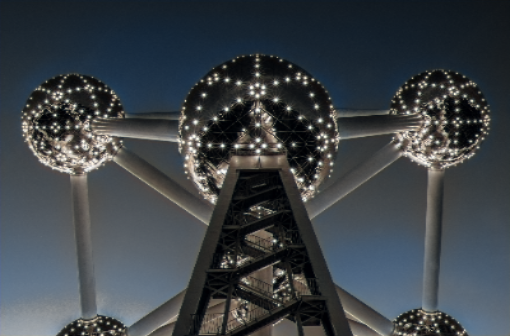} &
\includegraphics[ width=\xlinewidth\linewidth, height=\ylinewidth\linewidth]{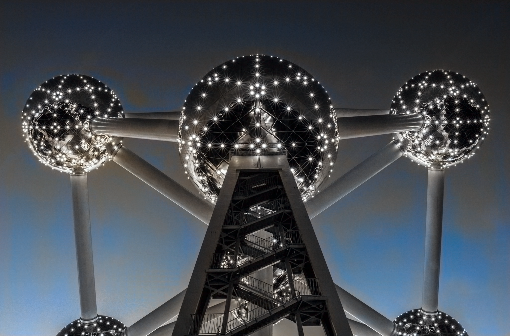} &
\includegraphics[ width=\xlinewidth\linewidth, height=\ylinewidth\linewidth]{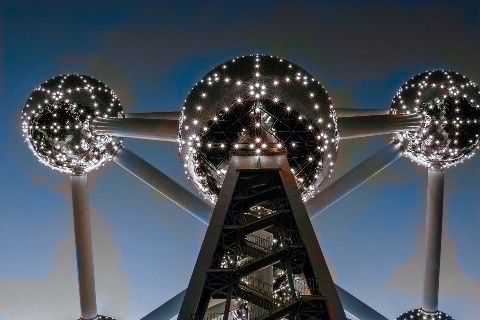} &
\includegraphics[ width=\xlinewidth\linewidth, keepaspectratio]{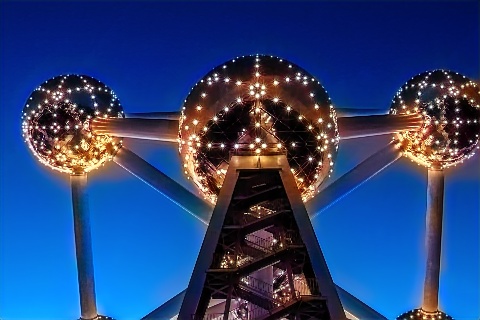}  \\[\xem]

\includegraphics[ width=\xlinewidth\linewidth, keepaspectratio]{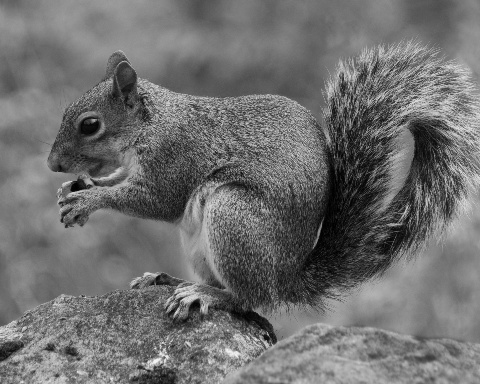} &
\includegraphics[ width=\xlinewidth\linewidth]{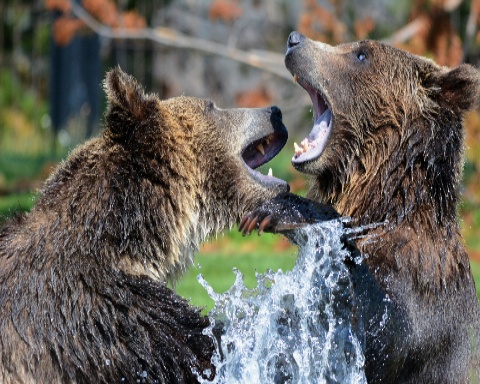} &
\includegraphics[ width=\xlinewidth\linewidth, height=0.108\linewidth]{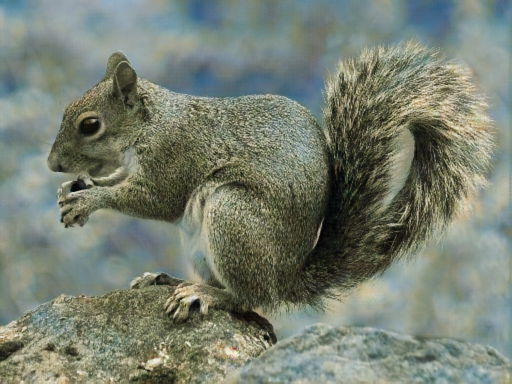} &
\includegraphics[ width=\xlinewidth\linewidth, height=0.108\linewidth]{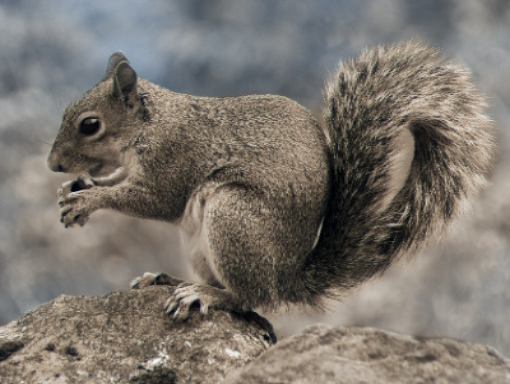} &
\includegraphics[ width=\xlinewidth\linewidth, height=0.108\linewidth]{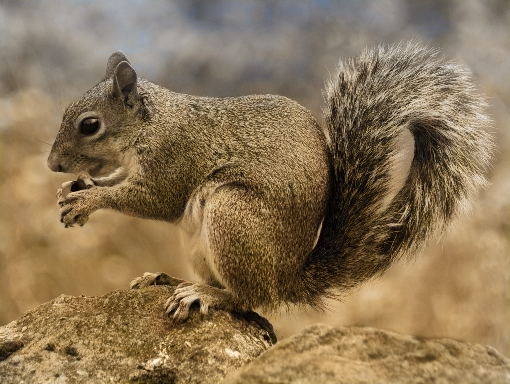} &
\includegraphics[ width=\xlinewidth\linewidth, keepaspectratio]{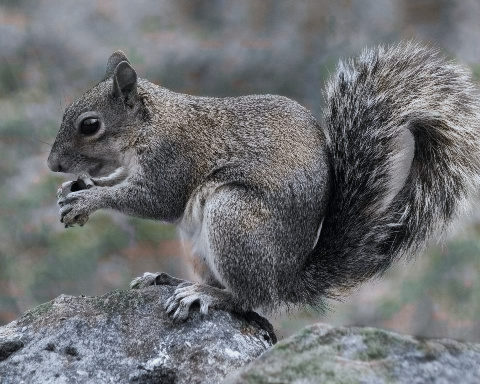} &
\includegraphics[ width=\xlinewidth\linewidth, keepaspectratio]{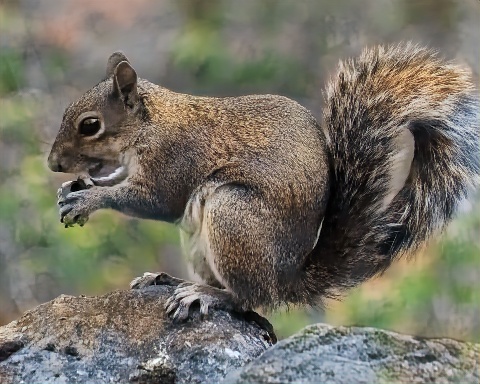}  \\[\xem]

\includegraphics[ width=\xlinewidth\linewidth, , height=\ylinewidth\linewidth]{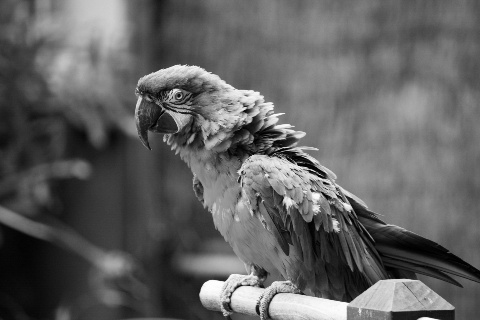} &
\includegraphics[ width=\xlinewidth\linewidth, height=\ylinewidth\linewidth]{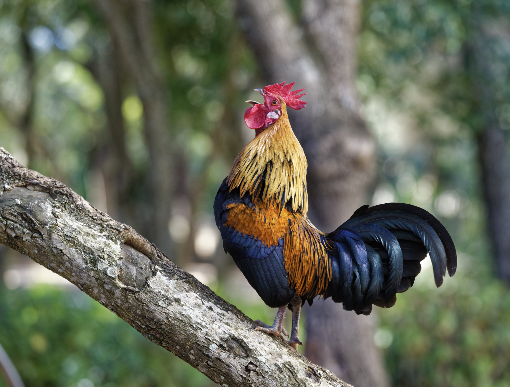} &
\includegraphics[ width=\xlinewidth\linewidth, , height=\ylinewidth\linewidth]{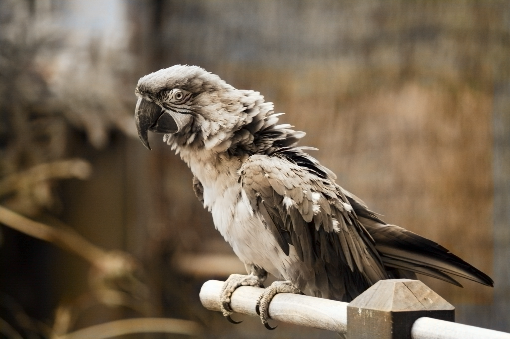} &
\includegraphics[ width=\xlinewidth\linewidth, , height=\ylinewidth\linewidth]{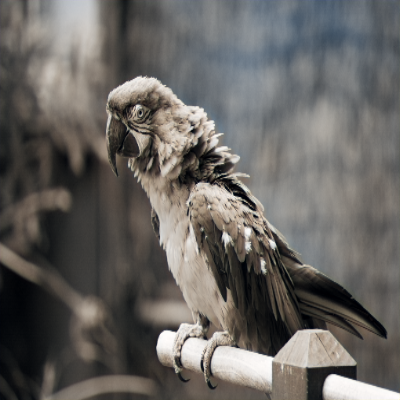} &
\includegraphics[ width=\xlinewidth\linewidth, , height=\ylinewidth\linewidth]{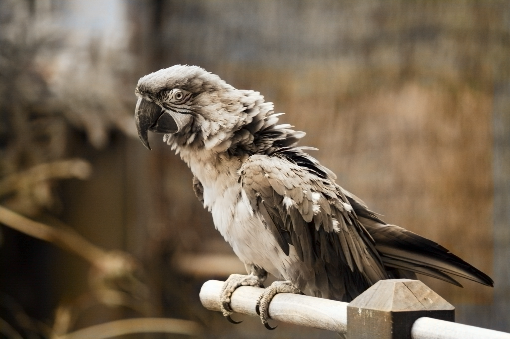} &
\includegraphics[ width=\xlinewidth\linewidth, , height=\ylinewidth\linewidth]{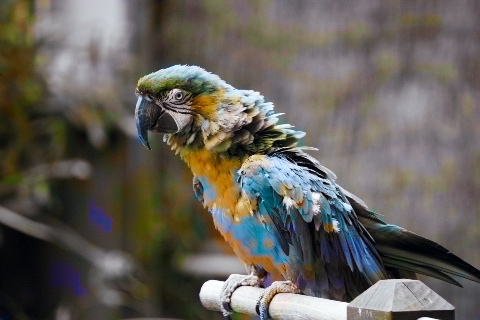} &
\includegraphics[ width=\xlinewidth\linewidth, , height=\ylinewidth\linewidth]{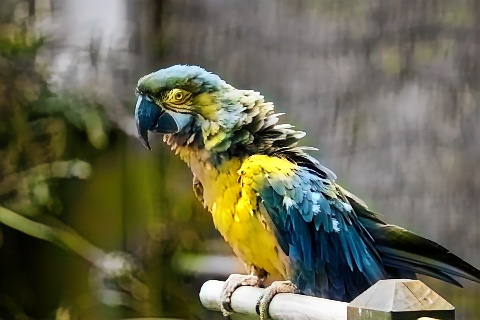}  \\[\xem]

Input & Reference & Pix2pix \cite{wang2018high}  & Deoldify \cite{jantic} & InstColor \cite{su2020instance} & He et al. \cite{he2018deep} & Ours \\

\end{tabular}

\caption{Visual comparisons against state-of-the-art colorization methods on DIV2K. \textmd{It shows that with only 800 training images, our method is able to accomplish visually pleasant colorization and our result is significantly better than others.}}
\label{fig:div2k_compare}
\vspace{-3mm}
\end{figure*}

\subsection{Training Objective}
\label{ssec:loss-fn}

In order to 1) simultaneously train the restoration and colorization nets, 2) exploit the rich color information available in the reference pictures, and 3) improve the visual quality of the overall restored output, we employed a weighted sum of diverse objectives functions against which the entire Pik-Fix system can be trained end-to-end. Among these, the \textit{luminance  reconstruction loss} between the restored luminances $I_L^{'}$ and the ground truth luminances $G_L$ are used to supervise the training of restoration subnet:
$\mathcal{L}_{rec,L}=||I_L^{'}-G_L||_1.$

However, it is well-known that relying on $\ell_p$ norms as loss function tends to generate blurred estimates of picture restoration~\cite{ledig2017photo}. Hence, we also used a measure of \textit{perceptual loss} that has been shown to deliver better quality visual results on a variety of restoration tasks~\cite{ledig2017photo, ding2021comparison, wang2018esrgan}:
$\mathcal{L}_{perc,L}=\sum_{j}\frac{1}{C_jH_jW_j} ||\phi_j(I_L^{'})-\phi_j(G_L)||^2_2,$
where $\phi_j$ is a feature map of shape $C_j\times H_j\times W_j$. 


The colorization subnet is intended to transfer color distributions from the reference picture to the predicted output pictures. Thus, we also use the \textit{histogram loss} to measure the distribution distance between the color histograms of the output and reference pictures as expressed by the Earth Mover's Distance (EMD):
$\mathcal{L}_\mathrm{EMD,\hbar}=\sum_{k=1}^K(\mathrm{CDF}_{\hbar_{I^{'}}}(k)-\mathrm{CDF}_{\hbar_{R}}(k))^2,$
where $\mathrm{CDF}_{p}(k)$ is the $k$-th element of the cumulative density function of the probability mass function $p$.
$\hbar_{I^{'}}$ and $\hbar_{R}$ are one-dimensional differentiable histograms formed by globally pooling over the SPHist features $h_{I^{'}}$ and $h_{R}$ in Eq.~\eqref{eq:histogra}, respectively.

We also use the \textit{chroma reconstruction loss} to impose the spatial consistency between the predicted chromatic channels $ab$ and the ground truth $ab$ channels, supplementing the histogram loss by directly controlling the pixel-wise chromatic loss:
$\mathcal{L}_{rec,ab}=||I_{ab}^{'}-G_{ab}||_1.$

The \textit{adversarial loss} is a recipe that is often used to enhance the visual quality of images synthesized using GANs~\cite{ledig2017photo, jiang2021enlightengan, kupyn2018deblurgan}. We utilize a PatchGAN~\cite{isola2017image} structure to ensure that all of the local patches of the enhanced output channels are visually similar to realistic chroma maps. The adversarial loss is expressed as:
$mathcal{L}_{adv,ab}=\mathbb{E}_{G_{ab}^{}}[\log{D(G)}]+\mathbb{E}_{I_{ab}^{'}}[\log{(1-D(I^{'},R))}].$

Finally, we combine all of the above directed loss functions into an overall loss under which Pik-Fix is trained: $\mathcal{L}= \alpha \mathcal{L}_{rec,L}+\beta\mathcal{L}_{perc,L}+  \lambda\mathcal{L}_{\mathrm{EMD},\hbar}+\gamma\mathcal{L}_{rec,ab}+\eta\mathcal{L}_{adv,ab}$.

\begin{table*}[!t]
\caption{\textbf{Quantitative comparison} on the DIV2K and Pascal VOC validation datasets. \textmd{Up-ward arrows indicate that a higher score denotes a good image quality. We highlight the best score for each measure.}}
\label{table:quan-result}
\setlength{\tabcolsep}{8pt}
\renewcommand{\arraystretch}{1.}
\centering
\footnotesize
\begin{tabular}{ lcccc cccc ccc}
\toprule
Dataset & \multicolumn{3}{c}{DIV2K (w/o degradation)} &  \multicolumn{3}{c}{Pascal VOC (w/o degradation)}
&  \multicolumn{3}{c}{Pascal VOC (w/ degradation)} 
\\ \cmidrule(lr){2-4}\cmidrule(
lr){5-7}
\cmidrule(lr){8-10}
Metric & PSNR$\uparrow$ & SSIM$\uparrow$ & LPIPS$\downarrow$ & PSNR$\uparrow$ & SSIM$\uparrow$ & LPIPS$\downarrow$ & PSNR$\uparrow$ & SSIM$\uparrow$ & LPIPS$\downarrow$ \\\midrule
Pix2pix  & 21.12 & 0.872 & 0.138 &  20.89 & 0.782 & 0.200  &20.37 & 0.732 & 0.231\\
DeOldify & 23.65 & 0.913 & 0.128 & 23.96 & 0.873 & 0.117 & 21.45 & 0.789 & 0.192 \\
He \textit{et al.}  & 23.53 & 0.918 & 0.125 &23.85 &0.925 &0.114 &- &- &-\\
InstColorization  & 22.45 & 0.914 & 0.131 & 23.95 & 0.932 & 0.111 & - & - & -  \\
Wan \textit{et al.}  - & - & - & - & - & - & - & 18.01 & 0.598 & 0.421 \\\midrule
Ours & \textbf{23.95} & \textbf{0.925} & \textbf{0.120} & \textbf{24.01} & \textbf{0.940} & \textbf{0.100} & \textbf{22.22} & \textbf{0.828} & \textbf{0.186} \\
\bottomrule
\end{tabular}
\vspace{-3mm}
\end{table*}

\subsection{Reference Picture Selection}
\label{ssec:ref-select}
As discussed earlier, Pik-Fix requires color reference pictures as additional inputs to guide the colorization process. Thus, we developed an automatic reference curation model that generates good reference pictures from a given database, given an input grayscale picture during either the training and or inferencing phases.

An ideal reference should be both visually and semantically similar to the target image to be colorized, while providing rich and appropriate color information for the colorizing process. Inspired by the deep perceptual similarity models~\cite{zhang2018unreasonable,ding2020image}, we leveraged a pre-trained VGG19 net~\cite{simonyan2014very} as backbone to extract intermediate deep feature maps. Then, we measured the degrees of textural and the structural similarity between each given grayscale input image and each of the available color images using the global means and variance/covariances of their feature maps, respectively~\cite{ding2020image}.
Finally, a weighted summation of the texture and structure similarities is used to determine which of the color reference pictures in the training set has the greatest similarity to the grayscale input picture to be repaired and colorized.
When deploying the trained Pik-Fik system, user may either automatically select a recommended reference picture retrieved from an available corpus, or they may choose to manually select a reference picture, according to their preference.

\begin{figure*}[!t]
\centering
\def\xlinewidth{0.12825}
\def\ylinewidth{0.855}
\def\ylineheight{0.168}
\def\ylineheighta{0.166}
\def\ylineheightb{0.13965}
\def\ylineheightc{0.168}
\def\xem{-1pt}
\def\yem{1pt}
\setlength{\tabcolsep}{1.2pt}
\begin{tabular}{ccccccc}

\includegraphics[ width=\xlinewidth\linewidth, height=\ylineheight\linewidth]{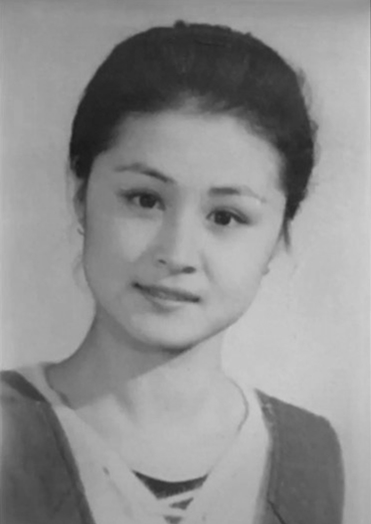} &
\includegraphics[ width=\xlinewidth\linewidth, height=\ylineheight\linewidth]{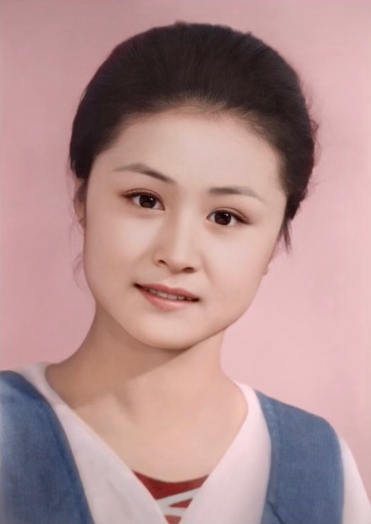} &
\includegraphics[ width=\xlinewidth\linewidth, height=\ylineheight\linewidth]{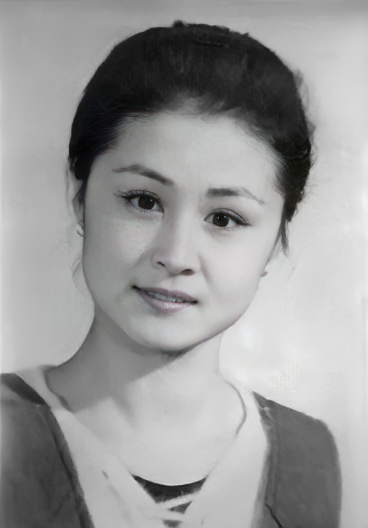} &
\includegraphics[ width=\xlinewidth\linewidth, height=\ylineheight\linewidth]{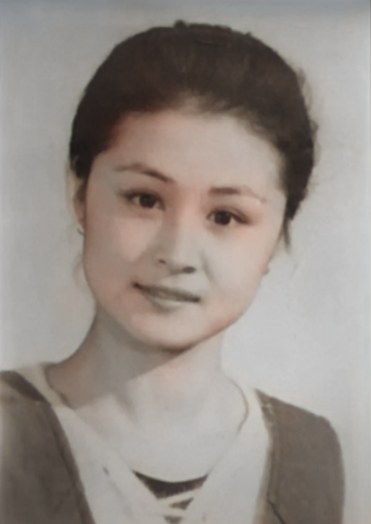} &
\includegraphics[ width=\xlinewidth\linewidth, height=\ylineheight\linewidth]{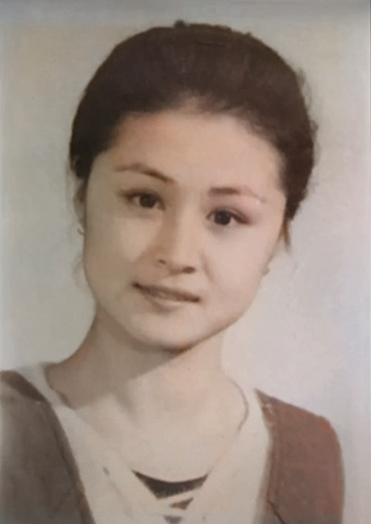} &
\includegraphics[ width=\xlinewidth\linewidth,height=\ylineheight\linewidth]{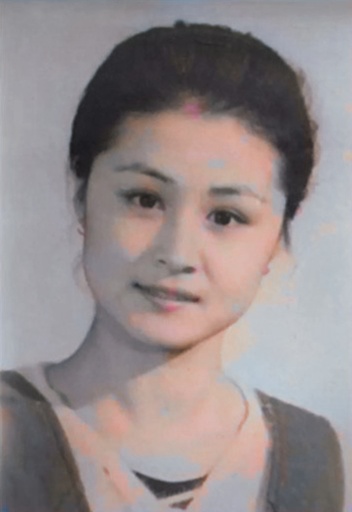} &
\includegraphics[ width=\xlinewidth\linewidth, height=\ylineheight\linewidth]{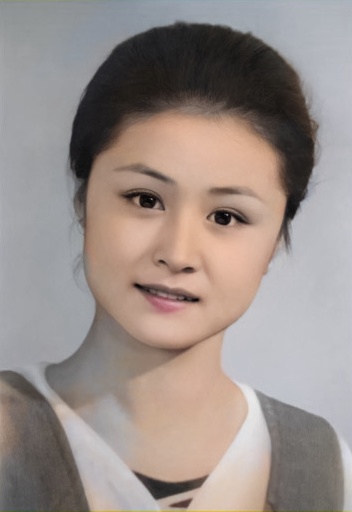} \\[\xem]

\includegraphics[ width=\xlinewidth\linewidth, height=\ylineheightc\linewidth]{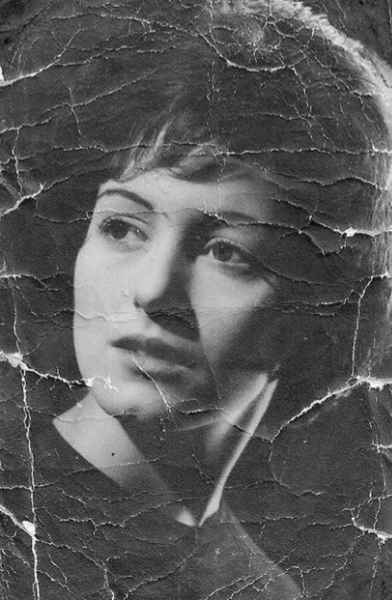} &
\includegraphics[ width=\xlinewidth\linewidth,height=\ylineheightc\linewidth]{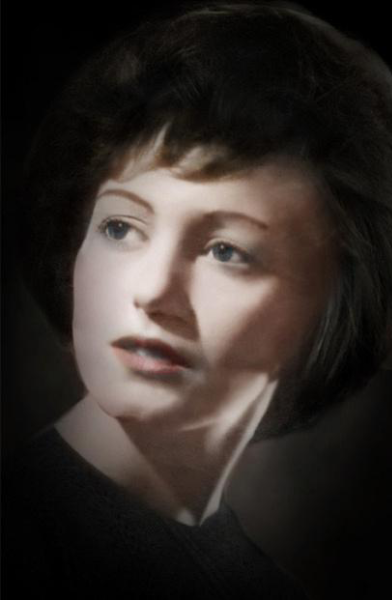} &
\includegraphics[ width=\xlinewidth\linewidth,height=\ylineheightc\linewidth]{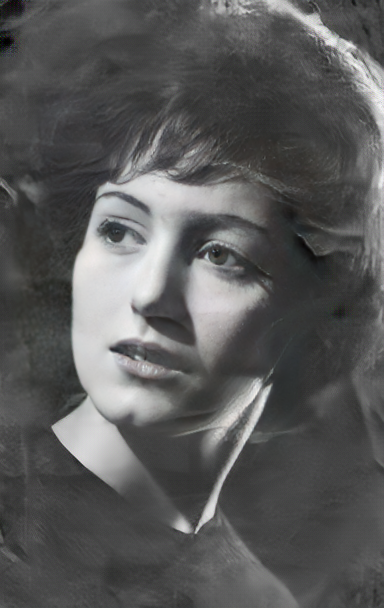} &
\includegraphics[ width=\xlinewidth\linewidth, height=\ylineheightc\linewidth]{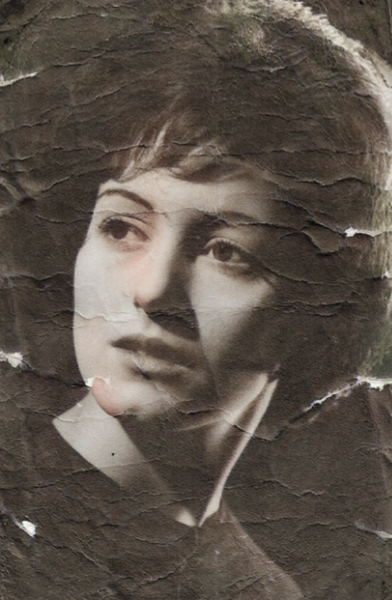} &
\includegraphics[ width=\xlinewidth\linewidth, height=\ylineheightc\linewidth]{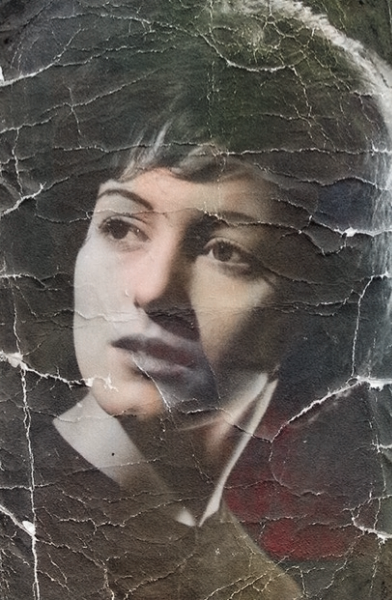} &
\includegraphics[ width=\xlinewidth\linewidth, height=\ylineheightc\linewidth]{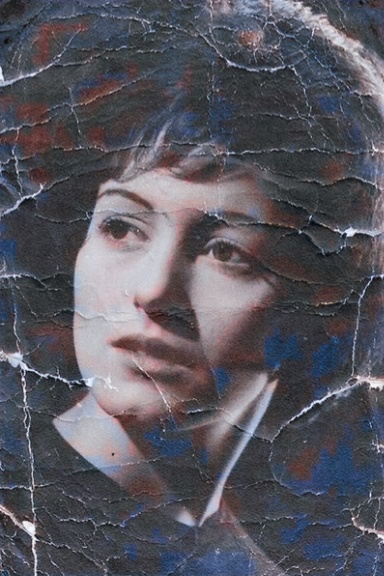} &
\includegraphics[ width=\xlinewidth\linewidth, height=\ylineheightc\linewidth]{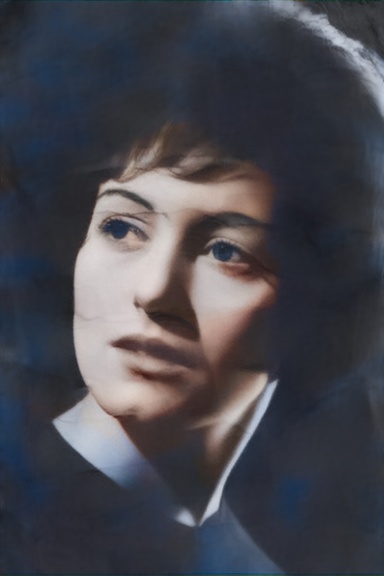} \\[\xem]

\includegraphics[ width=\xlinewidth\linewidth, , height=\ylineheighta\linewidth]{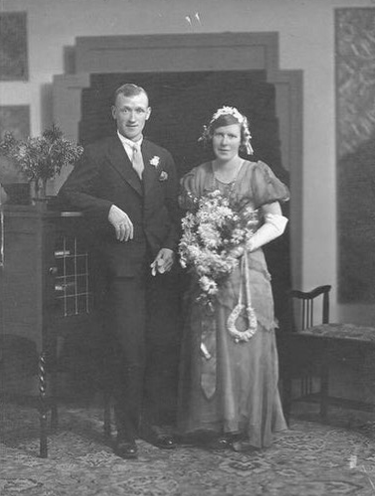} &
\includegraphics[ width=\xlinewidth\linewidth, height=\ylineheighta\linewidth]{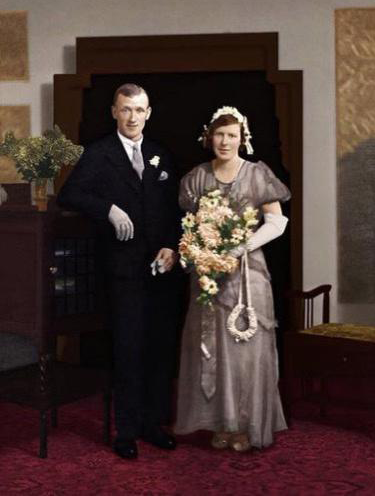} &
\includegraphics[ width=\xlinewidth\linewidth,height=\ylineheighta\linewidth]{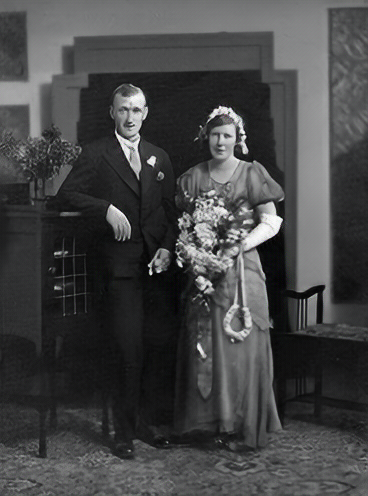} &
\includegraphics[ width=\xlinewidth\linewidth, height=\ylineheighta\linewidth]{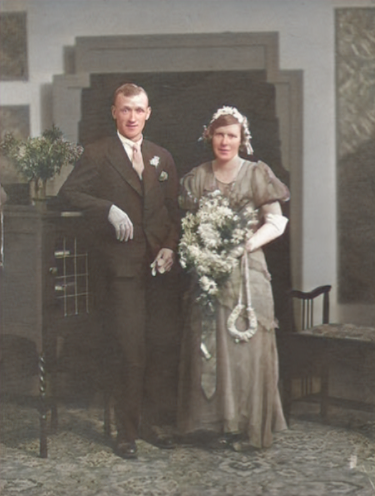} &
\includegraphics[ width=\xlinewidth\linewidth, height=\ylineheighta\linewidth]{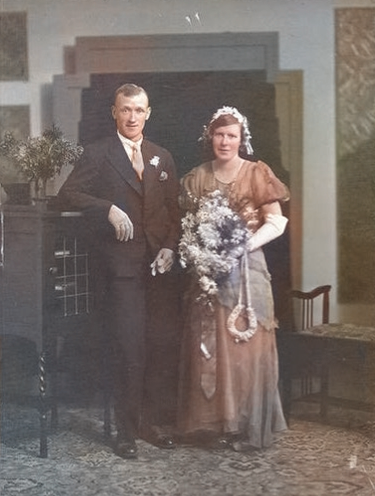} &
\includegraphics[ width=\xlinewidth\linewidth, height=\ylineheighta\linewidth]{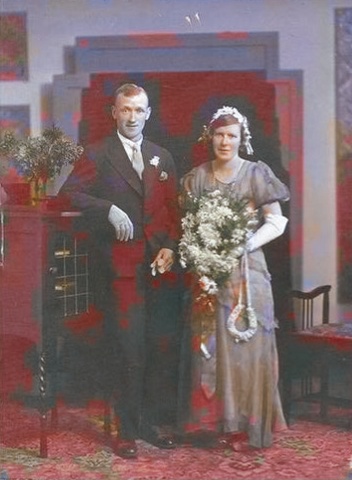} &
\includegraphics[ width=\xlinewidth\linewidth, height=\ylineheighta\linewidth]{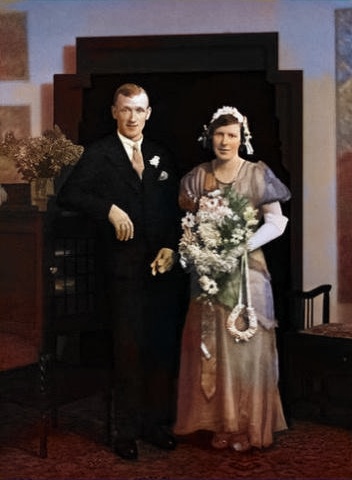} \\[\xem]

\includegraphics[ width=\xlinewidth\linewidth, , height=\ylineheightb\linewidth]{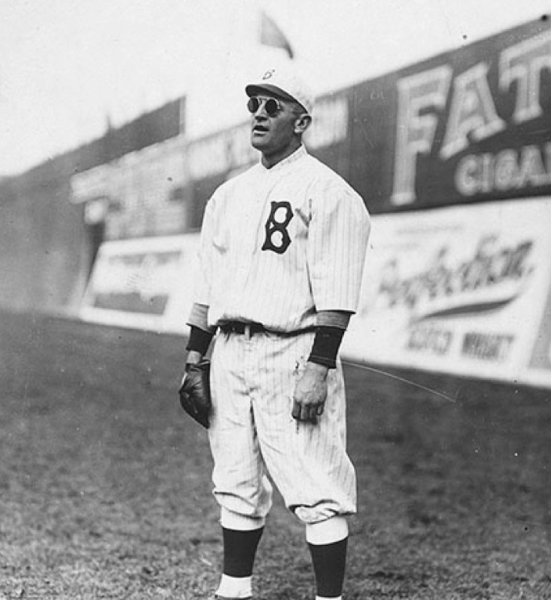} &
\includegraphics[ width=\xlinewidth\linewidth, height=\ylineheightb\linewidth]{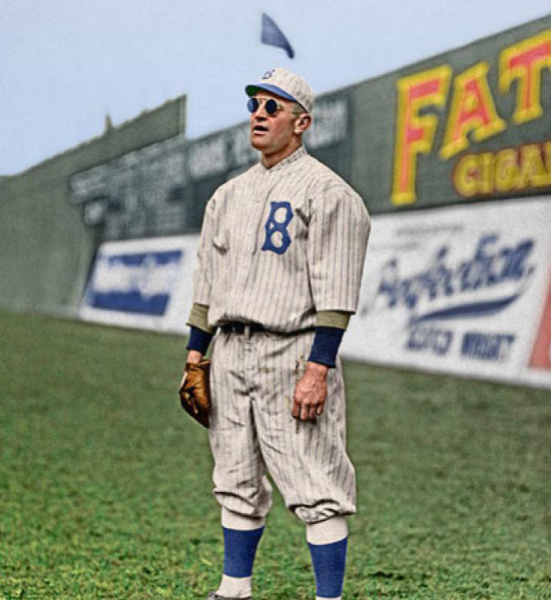} &
\includegraphics[ width=\xlinewidth\linewidth, height=\ylineheightb\linewidth]{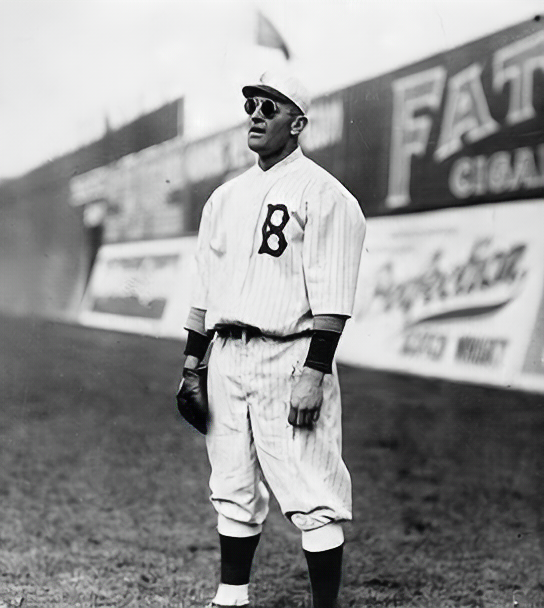} &
\includegraphics[ width=\xlinewidth\linewidth, height=\ylineheightb\linewidth]{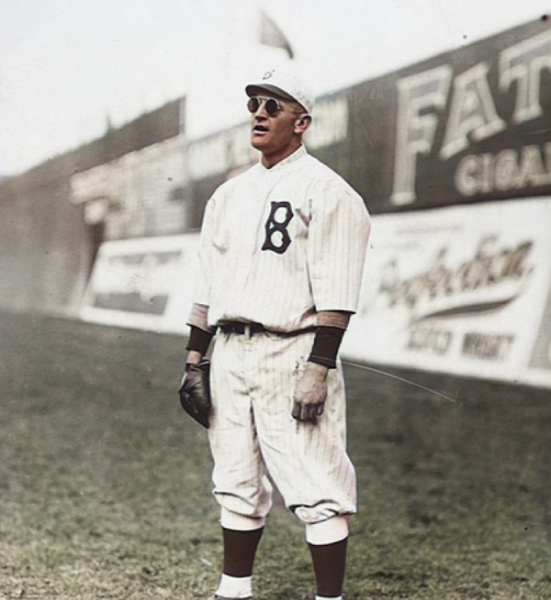} &
\includegraphics[ width=\xlinewidth\linewidth, height=\ylineheightb\linewidth]{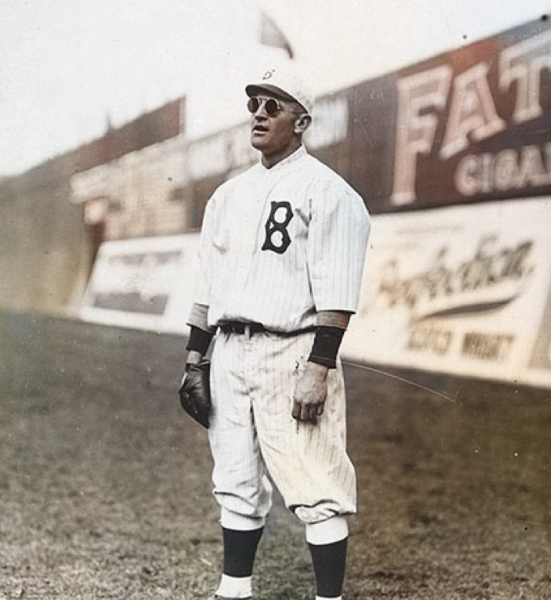} &
\includegraphics[ width=\xlinewidth\linewidth, height=\ylineheightb\linewidth]{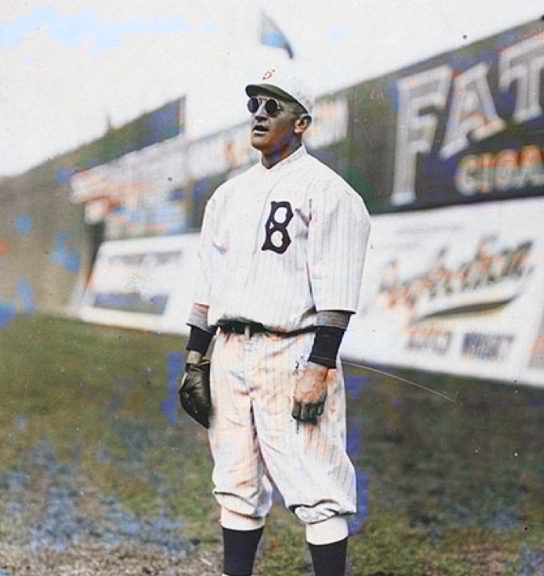} &
\includegraphics[ width=\xlinewidth\linewidth, height=\ylineheightb\linewidth]{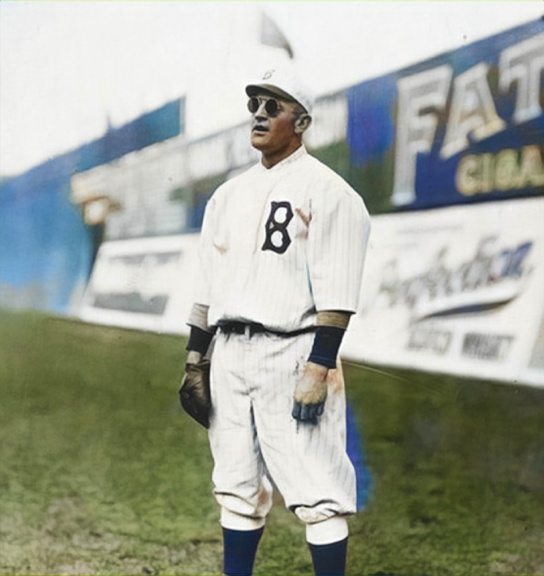} \\[\xem]

Input & Expert Repair & Wan et al. \cite{wan2020bringing}  & Deoldify \cite{jantic} & InstColor \cite{su2020instance} & He et al. \cite{he2018deep} & Ours \\

\end{tabular}
\caption{Visual comparisons against state-of-the-art colorization and  restoration methods on RealOld. \textmd{It shows that with limited synthetic training data from Pascal, our model is able to fix most of the degradation and deliver plausible colorization.}}
\label{fig:realold_compare}
\vspace{-4mm}
\end{figure*}

\section{Experiments}
\label{sec:experiments}
\subsection{Experimental Setting}
\textbf{\textup{Dataset}}
We trained and evaluated our method on three datasets: Div2K~\cite{Agustsson_2017_CVPR_Workshops}, Pascal~\cite{pascal}, and RealOld. 
In our experiments, we used the Div2K training and validation sets (800/100) for model training and testing, respectively.
For Pascal, we randomly selected 10,000/1000 images to serve as training data and testing data.
Two different experiments were conducted: simultaneous image restoration and colorization, and only image colorization.
 In order to produce realistic defect pictures, similar to those used in \cite{wan2020bringing}, we hired Photoshop experts to mimic the degradation patterns in real old photos (but not from our newly created RealOld dataset) on images from the Pascal dataset, using alpha compositing with randomized transparency levels, thereby generating synthetic old photos. We also added Gaussian blur, and simulated severe photo damage by randomly setting polygonal picture regions to pure white.
We restrict that reference images can only be retrieved from the training set.

\textit{Real-World Old Photos (RealOld)}:
To validate the efficacy and generalizability of our model under realistic conditions, we collected digitized copies of 200 real old black \& white photographs. Each of these photos were digitally manually restored and colorized by Photoshop experts. To the best of our knowledge, this is the first real-world old photo dataset that has aligned ``ground truth'' `pristine' photos to enable pixel-to-pixel processing and comparison.
We are making this dataset publicly available to allow other researchers to develop advanced algorithms that can both colorize and repair old photos impaired by scratches, blur, cracks, wear, film grain noise, and physical and capture distortions.
In our experiments, RealOld is used for testing models that have been trained on Pascal.
Furthermore, we randomly downloaded 2,000 RGB portraits from Google Images, and utilize our picture selection algorithm to pick the best references among them, when testing on ReadOld.

\noindent\textbf{\textup{Evaluation Metrics.}}
We report PSNR and SSIM~\cite{wang2004image} scores between the reference and restored/colorized pictures. As an alternative, we also use the learned perceptual image patch similarity (LPIPS) metric~\cite{zhang2018perceptual}. 


\noindent\textbf{\textup{Training Details.}}
We trained Pik-Fix in an end-to-end manner using the Adam solver~\cite{adam}, with $\beta_1=0.99$ and $\beta_2=0.999$. The initial learning rate was set to 0.0001 and exponentially decreased at the end of each epoch using a decay rate of 0.99. The loss balance weights were fixed as follows: $\alpha=1.0, \beta=0.2, \lambda=0.5, \gamma=1.0, \eta=0.2$. For data augmentation, randomly cropped $256 \times 256$ patches were included in the training data. At each epoch, we selected one of the following two methods of reference patch/picture generation: 1) a patch was cropped from an RGB image at a location different from that of the patch used as input, and processed with color jittering and a small affine transformation to create a reference picture; 2) one picture was randomly selected from the training set (excluding the ground-truth) and treated as the reference.  All of the compared models were trained on DIV2K and Pascal for 20 epochs, respectively, on a single GTX 3090Ti GPU.


\subsection{Experimental Results}

Since no existing work has explicitly considered the simultaneous correction of picture degradation and colorization, we compared Pik-Fix with models developed for image-to-image translation (denoted as Pix2pix~\cite{pix2pix2017}), image restoration (denoted as Wan \textit{et al.}~\cite{wan2020bringing}), and colorization (denoted as Deoldify~\cite{jantic}, He \textit{et al.}~\cite{he2018deep} and InstColorization~\cite{su2020instance}). For fair comparisons, we do not evaluate InstColorization and He \textit{et al.} (which do not restore degradation) on Pascal VOC with  degradation. Likewise, we did not compare against Wan \textit{et al.} (which does not colorize), on DIV2K without degradation or on Pascal VOC without degradation. All the compared methods were trained from scratch using the training strategies and code provided by those authors. 


\subsubsection{Quantitative Comparison}
Table~\ref{table:quan-result} compares Pik-Fix and the other models' performances on two public datasets under two scenarios: DIV2K without degradation, Pascal VOC without degradation, and Pascal VOC with degradation. Pik-Fix delivered the best results against all three evaluation metrics as compared with these state-of-the-art models. For example, on the DIV2K dataset without degradation, Pik-Fix achieved much better scores of 23.95 PSNR, 0.925 SSIM, and 0.120 LPIPS, than any of the compared models.

Table~\ref{table:quan-realold} shows the results obtained on the RealOld dataset, where again, Pik-Fix generated the highest performance scores among all compared models. These results strongly highlight the performance resilience by Pik-Fix when transferring from synthetic dataset to real-world old photo dataset.



\subsubsection{Qualitative Comparison}
Figure~\ref{fig:div2k_compare} provides qualitative comparison on Divk2K. Pik-Fix produced pictures having vivid, realistic colors, while compared models delivered incomplete colorization. Fig.~\ref{fig:realold_compare} shows results on the RealOld dataset, showing that Pik-Fix can simultaneously perform picture restoration and colorization, producing perceptually satisfying results.

\subsubsection{User Study}
We conducted a user study to compare the visual results of all the methods. We randomly selected 100 old photos from the RealOld dataset, and asked 15 users to rank the results based on their subjective visual impressions. We gathered reports from these 15 people with the results presented in Table~\ref{table:user_study}. Pik-Fix attained a high probability of 50.6\% of being selected as the single top performer, outperforming all of the other methods over all rankings, further illustrating the strong performance of the Pik-Fix picture restoration and colorization engine.


\subsection{Ablation Studies}

\noindent\textbf{Multi-scale SPHist.}
We conducted three experiments on the Div2k dataset to evaluate the effectiveness of multi-scale SPHist: 1) following~\cite{he2018deep}, the transferred $ab$ channels of the reference picture and the $L$ channel of the input are concatenated and fed into the backbone of the colorization sub-net; 2) instead of computing the multi-scale SPHist of the reference image, the multi-scale raw $ab$ channels of the reference image are used as input; 3) only a single-scale color histogram is fused with the shallower layers of the encoder. The results reported in Table~\ref{table:sphist_ablation} has validated the importance and usefulness of the proposed multi-scale SPHist.




\noindent\textbf{Multi-scale Similarity Maps.}
To validate the efficacy of multi-scale similarity maps relative to using a single similarity map, we conducted two experiments: 1) we use the single-scale similarity map proposed in~\cite{zhang2019deep}; 2) no similarity map is applied to the reference image. Table~\ref{table:multisim_ablation} reflects the benefits brought by the use of multi-scale similarity maps.

\begin{table}[!t]
\footnotesize
\caption{\textbf{Quantitative comparisons} of restoration/colorization performance on the RealOld dataset.}
\label{table:quan-realold}
\setlength{\tabcolsep}{6pt}
\renewcommand{\arraystretch}{0.95}
\centering
\begin{tabular}{ lccc}
\toprule
Dataset & \multicolumn{3}{c}{Real old photo}
\\ \cmidrule(lr){2-4}
Metric & PSNR$\uparrow$ & SSIM$\uparrow$ & LPIPS$\downarrow$ \\\midrule
Pix2pix  & 16.80 &0.684 &0.320 \\
DeOldify  & 17.14 & 0.723 & 0.287\\
He \textit{et al.}  &16.72 &0.707 &0.314  \\
InstColorization  & 16.86 & 0.715 & 0.312 \\
Wan \textit{et al.}  & 16.99 & 0.709 & 0.303 \\\midrule
Ours &  \textbf{17.20} & \textbf{0.758} & \textbf{0.258} \\
\bottomrule
\end{tabular}

\end{table}

\begin{table}[!t]
\footnotesize
\setlength{\tabcolsep}{6pt}
\renewcommand{\arraystretch}{1.}
\caption{\textbf{User rankings of algorithm performance} on the RealOld dataset. \textmd{The percentage (\%) of users choosing each model ranking is shown.}}
\label{table:user_study}
\begin{tabular}{llllll}
\toprule
Method           & Top 1         & Top 2         & Top 3         & Top 4         & Top 5         \\\midrule
Pix2Pix\cite{pix2pix2017}          & 3.1         & 11.8          & 24.3          & 41.6          & 67.7          \\
Deoldify\cite{jantic}         & 10.5          & 33.9          & 56.5          & 79.3          & 92.8          \\
He \textit{et al.} \cite{he2018deep}              & 4.6           & 26.3          & 44.1          & 61.9          & 83.2          \\
InstColorization \cite{su2020instance} & 8.0           & 26.3          & 53.6          & 75.5          & 92.2          \\
Wan \textit{et al.} \cite{wan2020bringing}              & 23.1          & 38.6          & 48.9          & 59.1          & 72.0          \\\midrule
\textbf{Ours}    & \textbf{50.6} & \textbf{65.9} & \textbf{75.3} & \textbf{85.1} & \textbf{94.4} \\ \bottomrule
\end{tabular}

\end{table}

\begin{figure}[!t]
\centering
\includegraphics[width=0.7\linewidth]{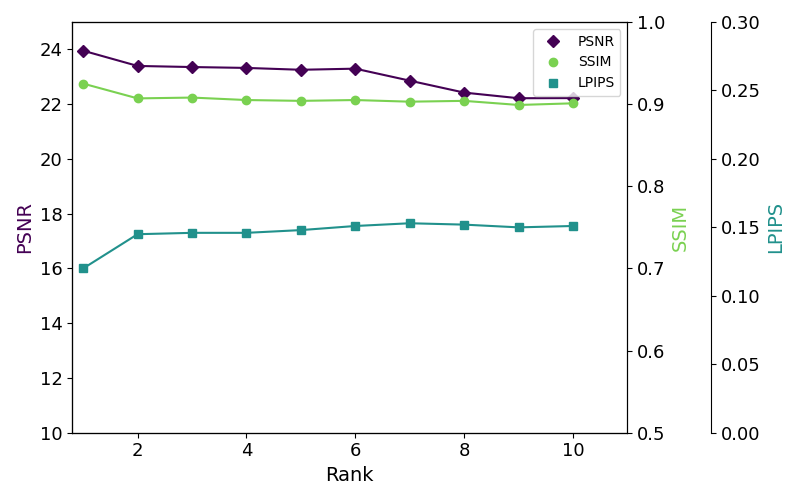}
\caption{Sensitivity analysis of reference image selection.}
\label{fig:reference_selection}
\end{figure}

\noindent\textbf{Multi-scale RDN.}
To study the possible performance gains brought by building a multi-level Residual Dense Network, we also tried the origin RDN~\cite{rdn} as the backbone for old photo restoration and tested the modified system on the Pascal dataset~\cite{pascal} with degradation. The results in Table~\ref{table:multirdn_ablation} show that the multi-level design significantly improve the quality of the restored outputs.

\begin{table}[!t]
\renewcommand{\arraystretch}{0.95}
\centering\footnotesize
\caption{Ablation study of multi-scale SPHist on Div2k.}
\label{table:sphist_ablation}
\begin{tabular}{llll}
\toprule
Method                        & PSNR$\uparrow$            & SSIM$\uparrow$           & LPIPS$\downarrow$          \\
\midrule
Input \textit{ab} fusion               & 22.978          & 0.902          & 0.130          \\
Multi-scale \textit{ab} fusion         & 23.233          & 0.910          & 0.127          \\
Single-scale histogram fusion & 23.631          & 0.906          & 0.125          \\
Multi-scale histogram fusion  & \textbf{23.952} & \textbf{0.925} & \textbf{0.120} \\
\bottomrule
\end{tabular}
\end{table}

\begin{table}[!t]
\renewcommand{\arraystretch}{0.95}
\centering\footnotesize
\caption{Ablation study of multi-scale similarity maps.}
\label{table:multisim_ablation}
\begin{tabular}{llll}
\toprule
Method                        & PSNR$\uparrow$            & SSIM$\uparrow$           & LPIPS$\downarrow$          \\
\midrule
No similarity map             & 22.817          & 0.910          & 0.131          \\
Single-scale similarity map   & 23.803          & 0.922          & 0.126         \\
Multi-scale similarity map   & \textbf{23.952} & \textbf{0.925} & \textbf{0.120} \\
\bottomrule
\end{tabular}

\end{table}

\begin{table}[!t]
\renewcommand{\arraystretch}{0.95}
\centering\footnotesize
\caption{Ablation study of multi-level RDN on Pascal.}
\label{table:multirdn_ablation}
\begin{tabular}{llll}
\toprule
Method                        & PSNR$\uparrow$            & SSIM$\uparrow$           & LPIPS$\downarrow$          \\
\midrule
Single-level RDN           & 21.89          & 0.818          & 0.190         \\
Multi-level RDN  & \textbf{22.22} & \textbf{0.828} & \textbf{0.186} \\
\bottomrule
\end{tabular}

\end{table}

\noindent\textbf{Sensitivity to Reference Image Selection.}
 To examine the robustness of our model relative to the selection of reference images, we compared the results obtained on the DIV2K dataset using different reference pictures than the computed ``most similar'' one (rank 1) to the least similar one (rank 10) among the ten selected reference pictures. As shown in Fig.~\ref{fig:reference_selection}, the performance of Pik-Fix is robust against reference picture selection, with only slight performance drops, \textit{viz.}, from 23.95 (rank 1) to 22.25 (rank 10) of PSNR, from 0.925 (rank 1) to 0.899 (rank 10) of SSIM, and from 0.12 (rank 1) to 0.15 (rank 10) of LPIPS.

\section{Concluding Remarks}
\label{sec:conclusion}
We propose a first end-to-end trainable system called Pik-Fix that is able to simultaneously restore and colorize old photos.
The overall system contains several subnetworks, each designed to handle a single defect, but trained holistically. A hierarchical restoration subnet recovers the luminance channel from physical and capture distortions, followed by a colorization subnet that uses space-preserving color histograms to estimate the chroma components. Extensive experimental results show that Pik-Fix attains excellent performance both visually and numerically on synthetic and real old photo datasets, as compared with state-of-the-art models. Moreover, we created the first publicly available real-world old photo dataset repaired by Photoshop experts, which we hope will facilitate further research on deep learning-based old photo restoration problems.

{\small
\bibliographystyle{ieee_fullname}
\bibliography{egbib}
}

\end{document}